\pdfoutput=1

\documentclass[final]{cvpr}

\usepackage{times}
\usepackage{epsfig}
\usepackage{graphicx}
\usepackage{amsmath}
\usepackage{amssymb}

\usepackage{float}
\usepackage{booktabs}
\usepackage{caption}
\usepackage{subcaption}
\usepackage{multirow}
\usepackage{xcolor}
\usepackage[utf8]{inputenc}

\usepackage[pagebackref=true,breaklinks=true,colorlinks,bookmarks=false]{hyperref}

\newcommand{\ba}{\mathbf{a}}\newcommand{\bA}{\mathbf{A}}

\newcommand{\bF}{\mathbf{F}} %

\newcommand{\bK}{\mathbf{K}}

\newcommand{\bM}{\mathbf{M}}

\newcommand{\bQ}{\mathbf{Q}}

\newcommand{\bV}{\mathbf{V}}
\newcommand{\bw}{\mathbf{w}}

\newcommand{\nE}{\mathbb{E}}

\newcommand{\nI}{\mathbb{I}}

\newcommand{\nR}{\mathbb{R}}

\newcommand{\cD}{\mathcal{D}}

\newcommand{\cG}{\mathcal{G}}

\newcommand{\cL}{\mathcal{L}}

\newcommand{\cW}{\mathcal{W}}
\newcommand{\cX}{\mathcal{X}}

\DeclareMathOperator*{\argmin}{argmin~}

\makeatletter
\DeclareRobustCommand\onedot{\futurelet\@let@token\@onedot}
\def\@onedot{\ifx\@let@token.\else.\null\fi\xspace}
\def\eg{e.g\onedot} 
\def\ie{i.e\onedot}

\makeatother

\newcommand{\boldparagraph}[1]{\vspace{0.1cm}\noindent{\bf #1:}}

\definecolor{darkgreen}{rgb}{0,0.7,0}
\definecolor{darkyellow}{rgb}{0.8,0.8,0}

\graphicspath{{gfx/}}

\begin{document}

\title{Multi-Modal Fusion Transformer for End-to-End Autonomous Driving}

\author{Aditya Prakash\thanks{indicates equal contribution} $^{1}$ \quad \quad Kashyap Chitta\footnotemark[1] $^{1,2}$ \quad \quad Andreas Geiger$^{1,2}$\\
$^{1}$Max Planck Institute for Intelligent Systems, T\"ubingen \quad \quad $^{2}$University of T\"ubingen\\
{\tt\small \{firstname.lastname\}@tue.mpg.de}
}

\maketitle

\begin{abstract}
How should representations from complementary sensors be integrated for autonomous driving? Geometry-based sensor fusion has shown great promise for perception tasks such as object detection and motion forecasting. However, for the actual driving task, the global context of the 3D scene is key, e.g. a change in traffic light state can affect the behavior of a vehicle geometrically distant from that traffic light. Geometry alone may therefore be insufficient for effectively fusing representations in end-to-end driving models. In this work, we demonstrate that imitation learning policies based on existing sensor fusion methods under-perform in the presence of a high density of dynamic agents and complex scenarios, which require global contextual reasoning, such as handling traffic oncoming from multiple directions at uncontrolled intersections. Therefore, we propose TransFuser, a novel Multi-Modal Fusion Transformer, to integrate image and LiDAR representations using attention. We experimentally validate the efficacy of our approach in urban settings involving complex scenarios using the CARLA urban driving simulator. Our approach achieves state-of-the-art driving performance while reducing collisions by 76\% compared to geometry-based fusion.
\end{abstract}
\section{Introduction}
\label{sec:intro}

\begin{figure}
    \centering
    \includegraphics[width=\columnwidth]{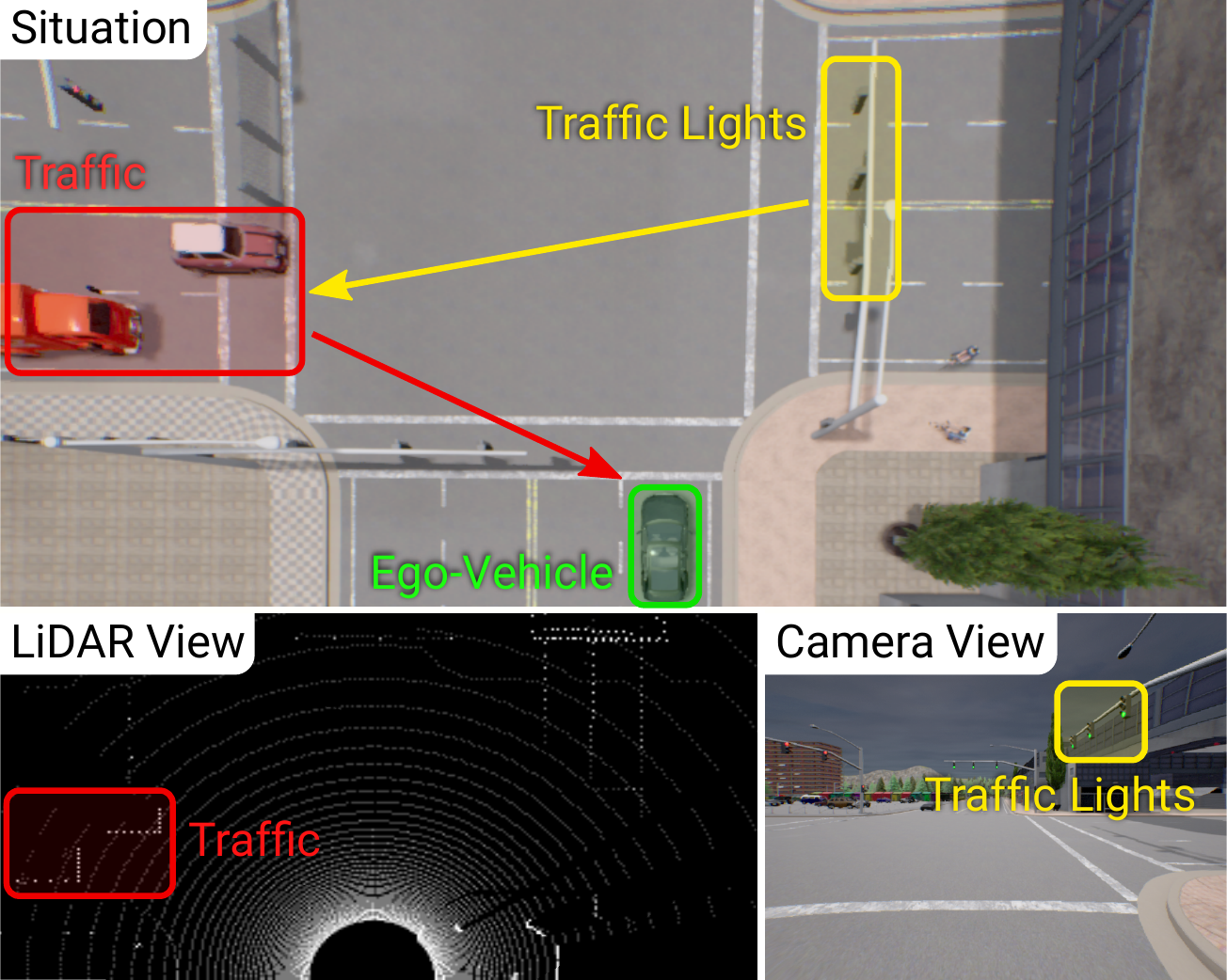}
    \caption{\textbf{Illustration.} Consider an intersection with oncoming traffic from the left. To safely navigate the intersection, the ego-vehicle ({\color{darkgreen}{green}}) must capture the global context of the scene involving the interaction between the traffic light ({\color{darkyellow}{yellow}}) and the vehicles ({\color{red}{red}}). However, the traffic light state is not visible in the LiDAR point cloud and the vehicles are not visible in the camera view. Our TransFuser model integrates both modalities via global attention mechanisms to capture the 3D context and navigate safely.}
    \label{fig:teaser}
    \vspace{-0.3cm}
\end{figure}

Image-only~\cite{Codevilla2019ICCV, Chen2019CORL, Ohn-Bar2020CVPR, Behl2020IROS, Prakash2020CVPR, Zhao2019ARXIV, Toromanoff2020CVPR} and LiDAR-only~\cite{Rhinehart2020ICLR, Filos2020ICML} methods have recently shown impressive results for end-to-end driving. However, these studies focus primarily on settings with limited dynamic agents and assume near-ideal behavior from other agents in the scene. With the introduction of \textit{adversarial scenarios} in the recent CARLA~\cite{Dosovitskiy2017CORL} versions, \eg vehicles running red lights, uncontrolled 4-way intersections, or pedestrians emerging from occluded regions to cross the road at random locations, image-only approaches perform unsatisfactory (\tabref{tab:results}) since they lack the 3D information of the scene required in these scenarios. While LiDAR consists of 3D information, LiDAR measurements are typically very sparse (in particular at distance), and additional sensors are required to capture information missing in LiDAR scans, \eg traffic light states.

While most existing methods for end-to-end driving focus on a single input modality, autonomous driving systems typically come equipped with both cameras and LiDAR sensors~\cite{Dosovitskiy2017CORL, Richter2016ECCV, Gaidon2016CVPR, Xu2017CVPRa, Cordts2016CVPR, Geiger2012CVPR, Ros2016CVPR, Waymo2019, Yu2018ARXIV}. This raises important questions: \textit{Can we integrate representations from these two modalities to exploit their complementary advantages for autonomous driving? To what extent should we process the different modalities independently and what kind of fusion mechanism should we employ for maximum performance gain?} Prior works in the field of {sensor fusion} have mostly focused on the perception aspect of driving, \eg 2D and 3D object detection~\cite{Fadadu2020ARXIV, Chen2017CVPR, Zhou2019CORL, Chen2020ARXIV, Qi2018CVPR, Ku2018IROSa, Liang2018ECCV, You2020ICLR, Liang2019CVPR, Meyer2019CVPRW}, motion forecasting~\cite{Fadadu2020ARXIV, Luo2018CVPR, Casas2020ICRA, Liang2020CVPR, Zhang2020CVPR, Casas2018CORL, Djuric2020ARXIV, Meyer2020ARXIV, Li2020IROS, Chen2020ARXIV}, and depth estimation~\cite{Fu2020ARXIV, Xu2019ICCV, You2020ICLR, Liang2019CVPR}. These methods focus on learning a state representation that captures the geometric and semantic information of the 3D scene. They operate primarily based on geometric feature projections between the image space and different LiDAR projection spaces, \eg Bird's Eye View (BEV)~\cite{Fadadu2020ARXIV, Chen2017CVPR, Zhou2019CORL, Chen2020ARXIV, Qi2018CVPR, Ku2018IROSa, Liang2018ECCV, You2020ICLR, Liang2019CVPR} and Range View (RV)~\cite{Meyer2019CVPR, Meyer2019CVPRW, Fadadu2020ARXIV, Meyer2020ARXIV, Chen2020ARXIV, Sobh2018NEURIPSW}. Information is typically aggregated from a local neighborhood around each feature in the projected 2D or 3D space. 

\begin{figure*}
\centering
\includegraphics[width=\textwidth]{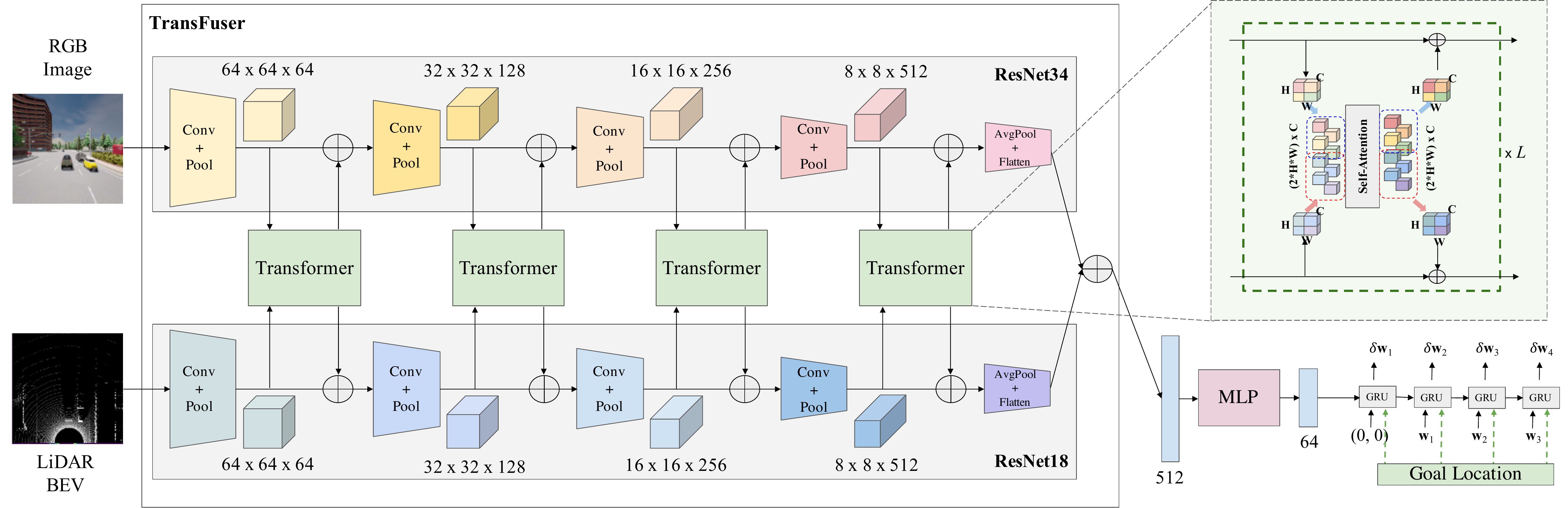}
\caption{\textbf{Architecture.} We consider single-view RGB image and LiDAR BEV representations (\secref{sec:io_parameterization}) as inputs to our Multi-Modal Fusion Transformer (TransFuser) which uses several transformer modules for the fusion of intermediate feature maps between both modalities. This fusion is applied at multiple resolutions ($64 \times 64$, $32 \times 32$, $16 \times 16$ and $8 \times 8$) throughout the feature extractor resulting in a 512-dimensional feature vector output from both the image and LiDAR BEV stream, which is combined via element-wise summation. This 512-dimensional feature vector constitutes a compact representation of the environment that encodes the global context of the 3D scene. It is then processed with an MLP before passing it to an auto-regressive waypoint prediction network. We use a single layer GRU followed by a linear layer which takes in the hidden state and predicts the differential ego-vehicle waypoints $\{ \delta \bw_{t}\}_{t=1}^T$, represented in the ego-vehicle's current coordinate frame.}
\label{fig:model}
\vspace{-0.2cm}
\end{figure*}

While these approaches fare better than image-only methods, we observe that the locality assumption in their architecture design hampers their performance in complex urban scenarios (\tabref{tab:driving_performance}). For example, when handling traffic at intersections, the ego-vehicle needs to account for interactions between multiple dynamic agents and traffic lights (\figref{fig:teaser}). While deep convolutional networks can be used to capture global context within a single modality, it is non-trivial to extend them to multiple modalities or model interactions between pairs of features. To overcome these limitations, we use the attention mechanism of transformers~\cite{Vaswani2017NEURIPS} to integrate global contextual reasoning about the 3D scene directly into the feature extraction layers of different modalities. We consider single-view image and LiDAR inputs since they are complementary to each other and our focus is on integrating representations from different types of modalities. We call the resulting model \emph{TransFuser} and integrate it into an auto-regressive waypoint prediction framework (\figref{fig:model}) designed for end-to-end driving.

\boldparagraph{Contributions} (1) We demonstrate that imitation learning policies based on existing sensor fusion approaches are unable to handle adversarial scenarios in urban driving, \eg, unprotected turnings at intersections or pedestrians emerging from occluded regions. (2) We propose a novel Multi-Modal Fusion Transformer (TransFuser) to incorporate the global context of the 3D scene into the feature extraction layers of different modalities. (3) We experimentally validate our approach in complex urban settings involving adversarial scenarios in CARLA and achieve state-of-the-art performance. Our code and trained models are available at \url{https://github.com/autonomousvision/transfuser}.

\section{Related Work}
\label{sec:related}

\boldparagraph{Multi-Modal Autonomous Driving} Recent multi-modal methods for end-to-end driving~\cite{Xiao2019ARXIV, Zhou2019SR, Sobh2018NEURIPSW, Behl2020IROS} have shown that complementing RGB images with depth and semantics has the potential to improve driving performance. Xiao et al.~\cite{Xiao2019ARXIV} explore RGBD input from the perspective of early, mid and late fusion of camera and depth modalities and observe significant gains. Behl et al.~\cite{Behl2020IROS} and Zhou et al.~\cite{Zhou2019SR} demonstrate the effectiveness of semantics and depth as explicit intermediate representations for driving. In this work, we focus on image and LiDAR inputs since they are complementary to each other in terms of representing the scene and are readily available in autonomous driving systems. In this respect, Sobh et al.~\cite{Sobh2018NEURIPSW} exploit a late fusion architecture for LiDAR and image modalities where each input is encoded in a separate stream and then concatenated together. However, we observe that this fusion mechanism suffers from high infraction rates in complex urban scenarios (\tabref{tab:infraction_analysis}) due to its inability to account for the behavior of multiple dynamic agents. Therefore, we propose a novel Multi-Modal Fusion Transformer that is effective in integrating information from different modalities at multiple stages during feature encoding and hence improves upon the limitations of the late fusion approach. 

\boldparagraph{Sensor Fusion Methods for Object Detection and Motion Forecasting} The majority of the sensor fusion works consider perception tasks, \eg object detection~\cite{Fadadu2020ARXIV, Chen2017CVPR, Zhou2019CORL, Chen2020ARXIVa, Qi2018CVPR, Ku2018IROSa, Liang2018ECCV, You2020ICLR, Liang2019CVPR, Meyer2019CVPRW} and motion forecasting~\cite{Luo2018CVPR, Casas2020ICRA, Liang2020CVPR, Zhang2020CVPR, Casas2018CORL, Djuric2020ARXIV, Meyer2020ARXIV}. They operate on multi-view LiDAR, \eg Bird's Eye View (BEV) and Range View (RV), or complement the camera input with depth information from LiDAR by projecting LiDAR features into the image space or projecting image features into the BEV or RV space. The closest approach to ours is ContFuse~\cite{Liang2018ECCV} which performs multi-scale dense feature fusion between image and LiDAR BEV features. For each pixel in the LiDAR BEV representation, it computes the nearest neighbors in a local neighborhood in 3D space, projects these neighboring points into the image space to obtain the corresponding image features, aggregates these features using continuous convolutions, and combines them with the LiDAR BEV features. Other projection-based fusion methods follow a similar trend and aggregate information from a local neighborhood in 2D or 3D space. However, the state representation learned by these methods is insufficient since they do not capture the global context of the 3D scene which is important for safe maneuvers in adversarial scenarios. To demonstrate this, we implement a multi-scale geometry-based fusion mechanism, inspired by~\cite{Liang2018ECCV, Liang2019CVPR}, involving both image-to-LiDAR and LiDAR-to-image feature fusion for end-to-end driving in CARLA and observe high infraction rates in the complex urban setting (\tabref{tab:infraction_analysis}). To overcome this limitation, we propose an attention-based Multi-Modal Fusion Transformer that incorporates global contextual reasoning and achieves superior driving performance.

\boldparagraph{Attention for Autonomous Driving} Attention has been explored in the context of driving for lane changing~\cite{Chen2019IROS}, object detection~\cite{Chen2017ARXIVa, Li2020IROS} and motion forecasting~\cite{Li2020IROS, Sadeghian2018ECCV, Sadeghian2019CVPR, Huang2019ICCV, Choi2019ICCV, Kosaraju2019NEURIPS, Ivanovic2019ICCV, Wei2020ARXIV}. Chen et al.~\cite{Chen2017ARXIVa} employ a recurrent attention mechanism over a learned semantic map for predicting vehicle controls. Li et al.~\cite{Li2020IROS} utilize attention to capture temporal and spatial dependencies between actors by incorporating a transformer module into a recurrent neural network. SA-NMP~\cite{Wei2020ARXIV} is a concurrent work that learns an attention mask over features extracted from a 2D CNN, operating on LiDAR BEV projections and HD maps, to focus on dynamic agents for safe motion planning. Chen et al.~\cite{Chen2019IROS} utilize attention in a hierarchical deep reinforcement learning framework to focus on the surrounding vehicles for lane changing in the TORCS racing simulator. They incorporate a spatial attention module to detect the most relevant regions in the image and a temporal attention module to weight different time-step image inputs, which leads to smoother lane changes. However, none of these approaches considers multiple modalities or encodes the global context of the 3D scene which is necessary for safely navigating adversarial scenarios. In contrast, we demonstrate the effectiveness of attention for feature fusion between different modalities on challenging urban driving scenarios.

\section{Method}
\label{sec:method}

In this work, we propose an architecture for end-to-end driving (\figref{fig:model}) with two main components: (1) a Multi-Modal Fusion Transformer for integrating information from multiple modalities (single-view image and LiDAR), and (2) an auto-regressive waypoint prediction network. The following sections detail our problem setting, input and output parameterizations, and each component of the model.

\subsection{Problem Setting} \label{sec:problem_setting}
We consider the task of point-to-point navigation in an urban setting~\cite{Filos2020ICML, Rhinehart2019ICCV, Rhinehart2020ICLR, Chen2019CORL, Codevilla2019ICCV} where the goal is to complete a given route while safely reacting to other dynamic agents and following traffic rules.

\boldparagraph{Imitation Learning (IL)} The goal of IL is to learn a policy $\pi$ that imitates the behavior of an expert $\pi^{*}$. In our setup, a policy is a mapping from inputs to waypoints that are provided to a separate low-level controller to output actions.
We consider the Behavior Cloning (BC) approach of IL which is a supervised learning method. An expert policy is first rolled out in the environment to collect a dataset, $\cD = \{ (\cX^i , \cW^i) \}_{i=1}^Z$ of size $Z$, which consists of high-dimensional observations of the environment, $\cX$, and the corresponding expert trajectory, defined by a set of 2D waypoints in BEV space, \ie, $\cW = \{ \bw_t = (x_t, y_t) \}_{t=1}^T$. This BEV space uses the coordinate frame of the ego-vehicle. The policy, $\pi$, is trained in a supervised manner using the collected data, $\cD$, with the loss function, $\cL$.
\begin{equation}
    \argmin_{\pi} \nE_{(\cX, \cW) \sim \cD} \left[ \cL (\cW, \pi(\cX)) \right]
\end{equation}
The high-dimensional observation, $\cX$, includes a front camera image input and a LiDAR point cloud from a single time-step. We use a single time-step input since prior works on IL for autonomous driving have shown that using observation histories may not lead to performance gain~\cite{Muller2005NEURIPS, Wang2019IROS, Bansal2019RSS, Wen2020NEURIPS}. We use the $L_1$ distance between the predicted trajectory, $\pi(\cX)$, and the expert trajectory, $\cW$, as the loss function. We assume access to an inverse dynamics model~\cite{Bellman2015Princeton}, implemented as a PID Controller $\nI$, which performs the low-level control, \ie, steer, throttle, and brake, provided the future trajectory $\cW$. The actions are determined as $\ba = \nI (\cW)$.

\boldparagraph{Global Planner} We follow the standard protocol of CARLA 0.9.10 and assume that high-level goal locations $\cG$ are provided as GPS coordinates. Note that these goal locations are sparse and can be hundreds of meters apart as opposed to the local waypoints predicted by the policy $\pi$.

\subsection{Input and Output Parameterization} \label{sec:io_parameterization}

\boldparagraph{Input Representation} Following ~\cite{Rhinehart2019ICCV, Filos2020ICML}, we convert the LiDAR point cloud into a 2-bin histogram over a 2D BEV grid with a fixed resolution. We consider the points within 32m in front of the ego-vehicle and 16m to each of the sides, thereby encompassing a BEV grid of 32m $\times$ 32m. We divide the grid into blocks of 0.125m $\times$ 0.125m which results in a resolution of 256 $\times$ 256 pixels. For the histogram, we discretize the height dimension into 2 bins representing the points on/below and above the ground plane. This results in a two-channel pseudo-image of size $256 \times 256$ pixels. For the RGB input, we consider the front camera with a FOV of 100$^{\circ}$. We extract the front image at a resolution of 400 $\times$ 300 pixels which we crop to 256 $\times$ 256 to remove radial distortion at the edges.

\boldparagraph{Output Representation} We predict the future trajectory $\cW$ of the ego-vehicle in BEV space, centered at the current coordinate frame of the ego-vehicle. The trajectory is represented by a sequence of 2D waypoints, $\{\bw_t = (x_t, y_t)\}_{t=1}^T$. We use $T=4$, which is the default number of waypoints required by our inverse dynamics model.

\subsection{Multi-Modal Fusion Transformer}

Our key idea is to exploit the self-attention mechanism of transformers~\cite{Vaswani2017NEURIPS} to incorporate the global context for image and LiDAR modalities given their complementary nature. The transformer architecture takes as input a sequence consisting of discrete tokens, each represented by a feature vector. The feature vector is supplemented by a positional encoding to incorporate positional inductive biases. 

Formally, we denote the input sequence as $\bF^{in} \in \nR^{N \times D_f}$, where $N$ is the number of tokens in the sequence and each token is represented by a feature vector of dimensionality $D_f$. The transformer uses linear projections for computing a set of queries, keys and values ($\bQ$, $\bK$ and $\bV$),
\begin{equation}
    \bQ = \bF^{in}\bM^q, \, \, \bK = \bF^{in}\bM^k, \, \, \bV = \bF^{in}\bM^v
\label{eqn:transformer}
\end{equation}
where $\bM^q \in \nR^{D_f \times D_q}$, $\bM^k \in \nR^{D_f \times D_k}$ and $\bM^v \in \nR^{D_f \times D_v}$ are weight matrices. It uses the scaled dot products between $\bQ$ and $\bK$ to compute the attention weights and then aggregates the values for each query,
\begin{equation}
    \bA = \text{softmax} \bigg(\frac{\bQ \bK^T} {\sqrt{D_k}}\bigg) \bV
\label{eqn:transformer_attention}
\end{equation}
Finally, the transformer uses a non-linear transformation to calculate the output features, $\bF^{out}$ which are of the same shape as the input features, $\bF^{in}$.
\begin{equation} \label{eqn:transformer_mlp}
    \bF^{out} = \text{MLP}(\bA) + \bF^{in}
\end{equation}
The transformer applies the attention mechanism multiple times throughout the architecture resulting in $L$ attention layers. Each layer in a standard transformer has multiple parallel attention `heads', which involve generating several $\bQ$, $\bK$ and $\bV$ values per $\bF^{in}$ for Eq. \eqref{eqn:transformer} and concatenating the resulting values of $\bA$ from Eq. \eqref{eqn:transformer_attention}.

Unlike the token input structures in NLP, we operate on grid structured feature maps. Similar to prior works on the application of transformers to images~\cite{Sun2019ICCV,Chen2020ICML,Qi2020ARXIV,Dosovitskiy2020ARXIV}, we consider the intermediate feature maps of each modality to be a set rather than a spatial grid and treat each element of the set as a token. The convolutional feature extractors for the image and LiDAR BEV inputs encode different aspects of the scene at different layers. Therefore, we fuse these features at multiple scales (\figref{fig:model}) throughout the encoder.

Let the intermediate grid structured feature map of a single modality be a 3D tensor of dimension $H \times W \times C$. For $S$ different modalities, these features are stacked together to form a sequence of dimension $(S*H*W) \times C$. We add a learnable positional embedding, which is a trainable parameter of dimension $(S*H*W) \times C$, so that the network can infer spatial dependencies between different tokens at train time. We also provide the current velocity as input by projecting the scalar value into a $C$ dimensional vector using a linear layer. The input sequence, positional embedding, and velocity embedding are combined using element-wise summation to form a tensor of dimension $(S*H*W) \times C$. As shown in \figref{fig:model}, this tensor is fed as input to the transformer which produces an output of the same dimension. We have omitted the positional embedding and velocity embedding inputs in~\figref{fig:model} for clarity. The output is then reshaped into $S$ feature maps of dimension $H \times W \times C$ each and fed back into each of the individual modality branches using an element-wise summation with the existing feature maps. The mechanism described above constitutes feature fusion at a single scale. This fusion is applied \textit{multiple times} throughout the ResNet feature extractors of the image and LiDAR BEV branches at different resolutions (\figref{fig:model}). However, processing feature maps at high spatial resolutions is computationally expensive. Therefore, we downsample higher resolution feature maps from the early encoder blocks using average pooling to a fixed resolution of $H=W=8$ before passing them as inputs to the transformer and upsample the output to the original resolution using bilinear interpolation before element-wise summation with the existing feature maps.

After carrying out dense feature fusion at multiple resolutions (\figref{fig:model}), we obtain a feature map of dimension $8 \times 8 \times 512$ from the feature extractors of each modality for an input of resolution $256 \times 256$ pixels. These feature maps are reduced to a dimension of $1 \times 1 \times 512$ by average pooling and flattened to a 512-dimensional feature vector. The feature vector of dimension 512 from both the image and the LiDAR BEV streams are then combined via element-wise summation. This 512-dimensional feature vector constitutes a compact representation of the environment that encodes the global context of the 3D scene. This is then fed to the waypoint prediction network which we describe next.

\subsection{Waypoint Prediction Network} As shown in~\figref{fig:model}, we pass the 512-dimensional feature vector through an MLP (comprising 2 hidden layers with 256 and 128 units) to reduce its dimensionality to 64 for computational efficiency before passing it to the auto-regressive waypoint network implemented using GRUs~\cite{Cho2014EMNLP}. We initialize the hidden state of the GRU with the 64-dimensional feature vector. The update gate of the GRU controls the flow of information encoded in the hidden state to the output and the next time-step. It also takes in the current position and the goal location (\secref{sec:problem_setting}) as input, which allows the network to focus on the relevant context in the hidden state for predicting the next waypoint. We provide the GPS coordinates of the goal location (registered to the ego-vehicle coordinate frame) as input to the GRU rather than the encoder since it lies in the same BEV space as the predicted waypoints and correlates better with them compared to representing the goal location in the perspective image domain~\cite{Chen2019CORL}. Following~\cite{Filos2020ICML}, we use a single layer GRU followed by a linear layer which takes in the hidden state and predicts the differential ego-vehicle waypoints $\{ \delta \bw_{t}\}_{t=1}^T$ for $T=4$ future time-steps in the ego-vehicle current coordinate frame. Therefore, the predicted future waypoints are given by $\{ \bw_t = \bw_{t-1} + \delta \bw_t \}_{t=1}^T$. The input to the first GRU unit is given as (0,0) since the BEV space is centered at the ego-vehicle's position.

\boldparagraph{Controller} We use two PID controllers for lateral and longitudinal control to obtain steer, throttle and brake values from the predicted waypoints, $\{ \bw_t \}_{t=1}^T$. The longitudinal controller takes in the magnitude of a weighted average of the vectors between waypoints of consecutive time-steps whereas the lateral controller takes in their orientation. For the PID controllers, we use the same configuration as in the author-provided codebase of~\cite{Chen2019CORL}. Implementation details can be found in the supplementary.

\subsection{Loss Function} Following~\cite{Chen2019CORL}, we train the network using an $L_1$ loss between the predicted waypoints and the ground truth waypoints (from the expert), registered to the current coordinate frame. Let $\bw_t^{gt}$ represent the ground truth waypoint for time-step $t$, then the loss function is given by:
\begin{equation}
\label{eqn:loss}
    \cL = \sum_{t=1}^{T} {||\bw_t - \bw_t^{gt}||}_1
\end{equation}
Note that the ground truth waypoints $\{\bw_t^{gt}\}$ which are available only at training time are different from the sparse goal locations $\cG$ provided at both training and test time.

\section{Experiments}
\label{sec:results}

In this section, we describe our experimental setup, compare the \textbf{driving performance} of our approach against several baselines, conduct an \textbf{infraction analysis} to study different failure cases, \textbf{visualize} the attention maps of TransFuser and present an \textbf{ablation study} to highlight the importance of different components of our model.

\boldparagraph{Task} \label{sec:task} We consider the task of navigation along a set of predefined routes in a variety of areas, \eg freeways, urban areas, and residential districts. The routes are defined by a sequence of sparse goal locations in GPS coordinates provided by a global planner and the corresponding discrete navigational commands, \eg follow lane, turn left/right, change lane. Our approach uses only the sparse GPS locations to drive. Each route consists of several scenarios, initialized at predefined positions, which test the ability of the agent to handle different kinds of adversarial situations, \eg obstacle avoidance, unprotected turns at intersections, vehicles running red lights, and pedestrians emerging from occluded regions to cross the road at random locations. The agent needs to complete the route within a specified time limit while following traffic regulations and coping with high densities of dynamic agents.

\boldparagraph{Dataset} We use the CARLA~\cite{Dosovitskiy2017CORL} simulator for training and testing, specifically CARLA 0.9.10 which consists of 8 publicly available towns. We use 7 towns for training and hold out Town05 for evaluation. For generating training data, we roll out an expert policy designed to drive using privileged information from the simulation and store data at 2FPS. Please refer to the supplementary material for additional details.
We select Town05 for evaluation due to the large diversity in drivable regions compared to other CARLA towns, \eg multi-lane and single-lane roads, highways and exits, bridges and underpasses. We consider two evaluation settings: (1) Town05 Short: 10 short routes of 100-500m comprising 3 intersections each, (2) Town05 Long: 10 long routes of 1000-2000m comprising 10 intersections each. Each route consists of a high density of dynamic agents and adversarial scenarios which are spawned at predefined positions along the route. Since we focus on handling dynamic agents and adversarial scenarios, we decouple this aspect from generalization across weather conditions and evaluate only on ClearNoon weather.

\boldparagraph{Metrics} We report results on 3 metrics. (1) \textbf{Route Completion (RC)}, percentage of route distance completed, (2) \textbf{Driving Score (DS)}, which is route completion weighted by an infraction multiplier that accounts for collisions with pedestrians, vehicles, and static elements, route deviations, lane infractions, running red lights, and running stop signs, and (3) \textbf{Infraction Count}. Additional details regarding the metrics and infractions are provided in the supplementary.

\boldparagraph{Baselines} We compare our TransFuser model to several baselines. (1) \textbf{CILRS}~\cite{Codevilla2019ICCV} is a conditional imitation learning method in which the agent learns to predict vehicle controls from a single front camera image while being conditioned on the navigational command. We closely follow the author-provided code and reimplement CILRS for CARLA 0.9.10 to account for the additional navigational commands compared to CARLA 0.8.4. (2) \textbf{LBC}~\cite{Chen2019CORL} is a knowledge distillation approach where a teacher model with access to ground truth BEV semantic maps is first trained using expert supervision to predict future waypoints followed by an image-based student model which is trained using supervision from the teacher. It is the current state-of-the-art approach on CARLA 0.9.6. We use the latest author-provided codebase for training on CARLA 0.9.10, which combines 3 input camera views by stacking different viewpoints as channels. (3) Auto-regressive IMage-based waypoint prediction \textbf{(AIM)}: We implement our auto-regressive waypoint prediction network with an image-based ResNet-34 encoder which takes just the front camera image as input. This baseline is equivalent to adapting the CILRS model to predict waypoints conditioned on sparse goal locations rather than vehicle controls conditioned on navigational commands. The image encoder used for this is the same as CILRS and our model. (4) \textbf{Late Fusion}: We implement a version of our architecture where the image and the LiDAR features are extracted independent of each other using the same encoders as TransFuser but without the transformers (similar to~\cite{Sobh2018NEURIPSW}), which are then fused through element-wise summation and passed to the waypoint prediction network. (5) \textbf{Geometric Fusion}: We implement a multi-scale geometry-based fusion method, inspired by~\cite{Liang2018ECCV, Liang2019CVPR}, involving both image-to-LiDAR and LiDAR-to-image feature fusion. We unproject each 0.125m $\times$ 0.125m block in our LiDAR BEV representation into 3D space resulting in a 3D volume. We randomly select 5 points from the LiDAR point cloud lying in this 3D volume and project them into the image space. We aggregate the image features of these points via element-wise summation before passing them to a 3-layer MLP. The output of the MLP is then combined with the LiDAR BEV feature of the corresponding 0.125m $\times$ 0.125m block at multiple resolutions throughout the feature extractor. Similarly, for each image pixel, we aggregate information from the LiDAR BEV features at multiple resolutions. This baseline is equivalent to replacing the transformers in our architecture with projection-based feature fusion.

We also report results for the expert used for generating our training data, which defines an upper bound for the performance on each evaluation setting. We provide additional details regarding all the baselines in the supplementary. 

\boldparagraph{Implementation Details} We use 2 sensor modalities, the front camera RGB image and LiDAR point cloud converted to BEV representation (\secref{sec:io_parameterization}), \ie, $S=2$. The RGB image is encoded using a ResNet-34~\cite{He2016CVPR} which is pre-trained on ImageNet~\cite{Deng2009CVPR}. The LiDAR BEV representation is encoded using a ResNet-18~\cite{He2016CVPR} which is trained from scratch. In our default TransFuser configuration, we use 1 transformer per resolution and 4 attention heads for each transformer. We select $D_q, D_k, D_v$ from $\{ 64, 128, 256, 512 \}$ for the 4 transformers corresponding to the feature embedding dimension $D_f$ at each resolution. For each of our baselines, we tested different perception backbone and chose the best: ResNet-34 for CILRS and AIM, ResNet-50 for LBC, ResNet-34 as the image encoder and ResNet-18 as the LiDAR BEV encoder for each of the sensor fusion methods. Additional details can be found in the supplementary.

\begin{table*}[t]
    \begin{tabular}{c c}
        \begin{subtable}[h]{0.58\textwidth}
            \small
            \setlength{\tabcolsep}{2.2pt}
            \centering
            \begin{tabular}{c | c c | c c }
                \textbf{Method} & \multicolumn{2}{c|}{\textbf{Town05 Short}} & \multicolumn{2}{c}{\textbf{Town05 Long}} \\
                \hline
                & DS $\uparrow$ & RC $\uparrow$ & DS $\uparrow$ & RC $\uparrow$ \\
                \hline
                CILRS~\cite{Codevilla2019ICCV} & 7.47 $\pm$ 2.51 & 13.40 $\pm$ 1.09 & 3.68 $\pm$ 2.16 & 7.19 $\pm$ 2.95 \\
                LBC~\cite{Chen2019CORL} & 30.97 $\pm$ 4.17 & 55.01 $\pm$ 5.14 & 7.05 $\pm$ 2.13 & 32.09 $\pm$ 7.40 \\
                AIM & 49.00 $\pm$ 6.83 & 81.07 $\pm$ 15.59 & 26.50 $\pm$ 4.82 & 60.66 $\pm$ 7.66 \\
                Late Fusion & 51.56 $\pm$ 5.24 & 83.66 $\pm$ 11.04 & 31.30 $\pm$ 5.53 & 68.05 $\pm$ 5.39 \\
                Geometric Fusion & 54.32 $\pm$ 4.85 & \textbf{86.91} $\pm$ 10.85 & 25.30 $\pm$ 4.08 & \textbf{69.17} $\pm$ 11.07 \\
                TransFuser (Ours) & \textbf{54.52} $\pm$ 4.29 & 78.41 $\pm$ 3.75 & \textbf{33.15} $\pm$ 4.04 & 56.36 $\pm$ 7.14 \\
                \hline
                \textit{Expert} & \textit{84.67 $\pm$ 6.21} & \textit{98.59 $\pm$ 2.17} & \textit{38.60 $\pm$ 4.00} & \textit{77.47 $\pm$ 1.86} \\
                \hline
            \end{tabular}
            \caption{\textbf{Driving Performance.} We report the mean and standard deviation over 9 runs of each method (3 training seeds, each seed evaluated 3 times) on 2 metrics: Route Completion (RC) and Driving Score (DS), in Town05 Short and Town05 Long settings comprising high densities of dynamic agents and scenarios.}
            \vspace{-0.2cm}
            \label{tab:driving_performance}
        \end{subtable} 
        & 
        \begin{subtable}[h]{0.36\textwidth}
            \includegraphics[width=\columnwidth]{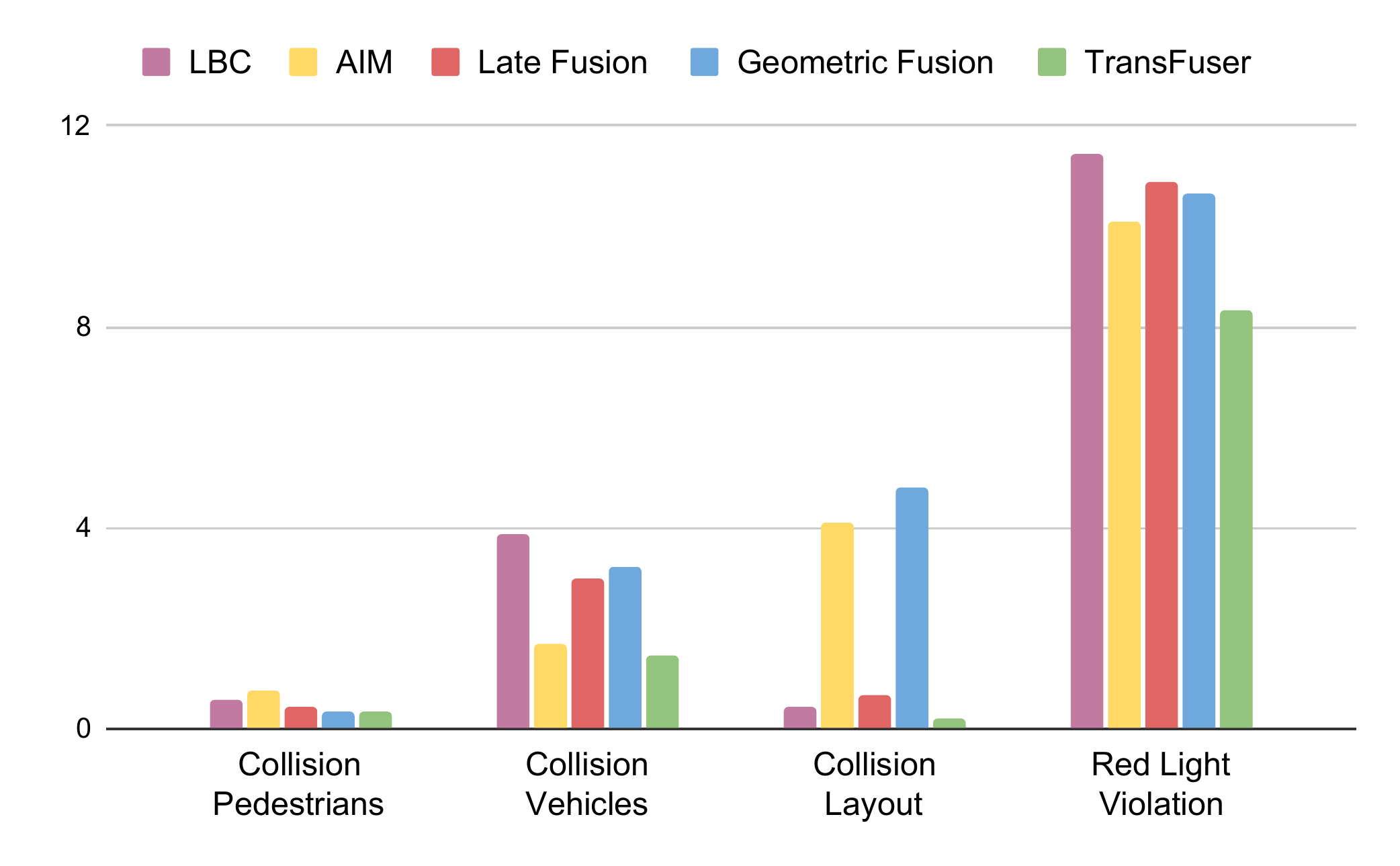}
            \caption{\textbf{Infractions.} We report the mean value of the total infractions incurred by each model over the 9 evaluation runs in the Town05 Short setting.}
            \vspace{-0.2cm}
            \label{tab:infraction_analysis}
        \end{subtable}
    \end{tabular}
    \caption{\textbf{Results.} We compare our TransFuser model with CILRS, LBC, auto-regressive image-based waypoint prediction network (AIM), and sensor fusion methods (Late Fusion of image and LiDAR features, Geometric feature projections between image and LiDAR BEV space) in terms of driving performance (\tabref{tab:driving_performance}) and infractions incurred (\tabref{tab:infraction_analysis}).}
    \label{tab:results}
    \vspace{-0.3cm}
\end{table*}

\subsection{Results}

\boldparagraph{Performance of CILRS and LBC} In our first experiment, we examine to what extent the current image-based methods on CARLA scale to the new 0.9.10 evaluation setting involving complex multi-lane intersections, adversarial scenarios, and heavy infraction penalties. From the results in~\tabref{tab:driving_performance} we observe that CILRS performs poorly on all evaluation settings. This is not surprising since CILRS is conditioned on discrete navigational commands whose data distribution is imbalanced as shown in the supplementary. While the original LBC~\cite{Chen2019CORL} architecture uses only the front camera image as input, the authors recently released an updated version of their architecture with 2 major modifications, (1) multi-view camera inputs (front, 45$^\circ$ left, and 45$^\circ$ right), (2) target heatmap as input (instead of navigational command) which is formed by projecting the sparse goal location in the image space. We train their updated model on our data and observe that LBC performs significantly better than CILRS on the short routes (\tabref{tab:driving_performance}), which is expected since it is trained using supervision from the teacher model which uses ground truth BEV semantic labels. However, LBC's performance drops drastically when evaluated on the long routes where it achieves 32.09 RC but suffers multiple infractions resulting in a low DS of 7.05. This is due to the frequent red light infractions and collision with vehicles (\tabref{tab:infraction_analysis}) resulting in large multiplicative penalties on the DS. These results show that CILRS and LBC are unable to handle the complexities of urban driving.

\boldparagraph{AIM is a strong baseline} Since the performance of CILRS and LBC drops significantly on the long routes, we focus on designing a strong image-based baseline next. Towards this goal, we replace the learned controller of CILRS with our auto-regressive waypoint prediction network. We observe that AIM significantly outperforms CILRS on all evaluation settings (\tabref{tab:driving_performance}), achieving nearly 7 times better performance. This is likely because AIM uses our inverse dynamics model (PID controller) for low-level control and represents goal locations in the same BEV coordinate space in which waypoints are predicted. In contrast, LBC's goal locations are represented as heatmaps in image space. Furthermore, AIM uses an auto-regressive GRU-based waypoint prediction network which enables the processing of these goal locations directly at the final stage of the network. This provides a prior that simplifies the learning of behaviors that follow the path to the goal location which could make the encoder prioritize information regarding high-level semantics of the scene, \eg traffic light state, rather than features relevant for low-level control. AIM outperforms LBC by 58.21\% on the short routes and 275.89\% on the long routes. The red light violations of LBC lead to a compounding of other infractions (\eg collisions with vehicles), which rapidly drops its DS compared to AIM.

\boldparagraph{Sensor Fusion Methods} The goal of this experiment is to determine the impact of the LiDAR modality on the driving performance and compare different fusion methods. For this, we compare our TransFuser to two baselines, Late Fusion (LF) and Geometric Fusion (GF). We observe that LF outperforms AIM on all evaluation settings (\tabref{tab:driving_performance}). This is expected since LiDAR provides additional 3D context which helps the agent to better navigate urban environments. Furthermore, we observe even better performance on the short routes when replacing the independent feature extractors of image and LiDAR branches with multi-scale geometry-based fusion encoder. However, both LF and GF suffer from a sharp drop in DS compared to their RC. We hypothesize that this occurs because they do not incorporate global contextual reasoning which is necessary to safely navigate the intersections, and focus primarily on navigation to the goal at all costs while ignoring obstacles which leads to several infractions~\tabref{tab:infraction_analysis}. This has a compounding effect on the long routes due to the exponential nature of the infraction penalty, resulting in a rapid drop in DS. In contrast, our TransFuser model outperforms GF by 31.02\% on DS with an 18.52\% lower RC on Town05 Long. It also achieves a 51.58\% reduction compared to LF and 76.11\% reduction compared to GF in collisions and 23.5\% reduction compared to LF and 21.93\% reduction compared to GF in red light violations. This shows that our model drives cautiously and focuses on dynamic agents and traffic lights. This indicates that attention is effective in incorporating the global context of the 3D scene which allows for safe driving. We provide driving videos in the supplementary.

\boldparagraph{Limitations} We observe that all fusion methods struggle with red light violations (\tabref{tab:infraction_analysis}). This is because detecting red lights is very challenging in Town05 since they are located on the opposite side of the intersection and are barely visible in the input image. Unlike some existing methods~\cite{Toromanoff2020CVPR}, we do not use any semantic supervision for red lights which furthers exacerbates this issue since the learning signal for red light detection is very weak. We expect the red light detection performance of the fusion approaches to improve when incorporating such additional supervision.

\begin{figure}
    \centering
    \includegraphics[width=\columnwidth]{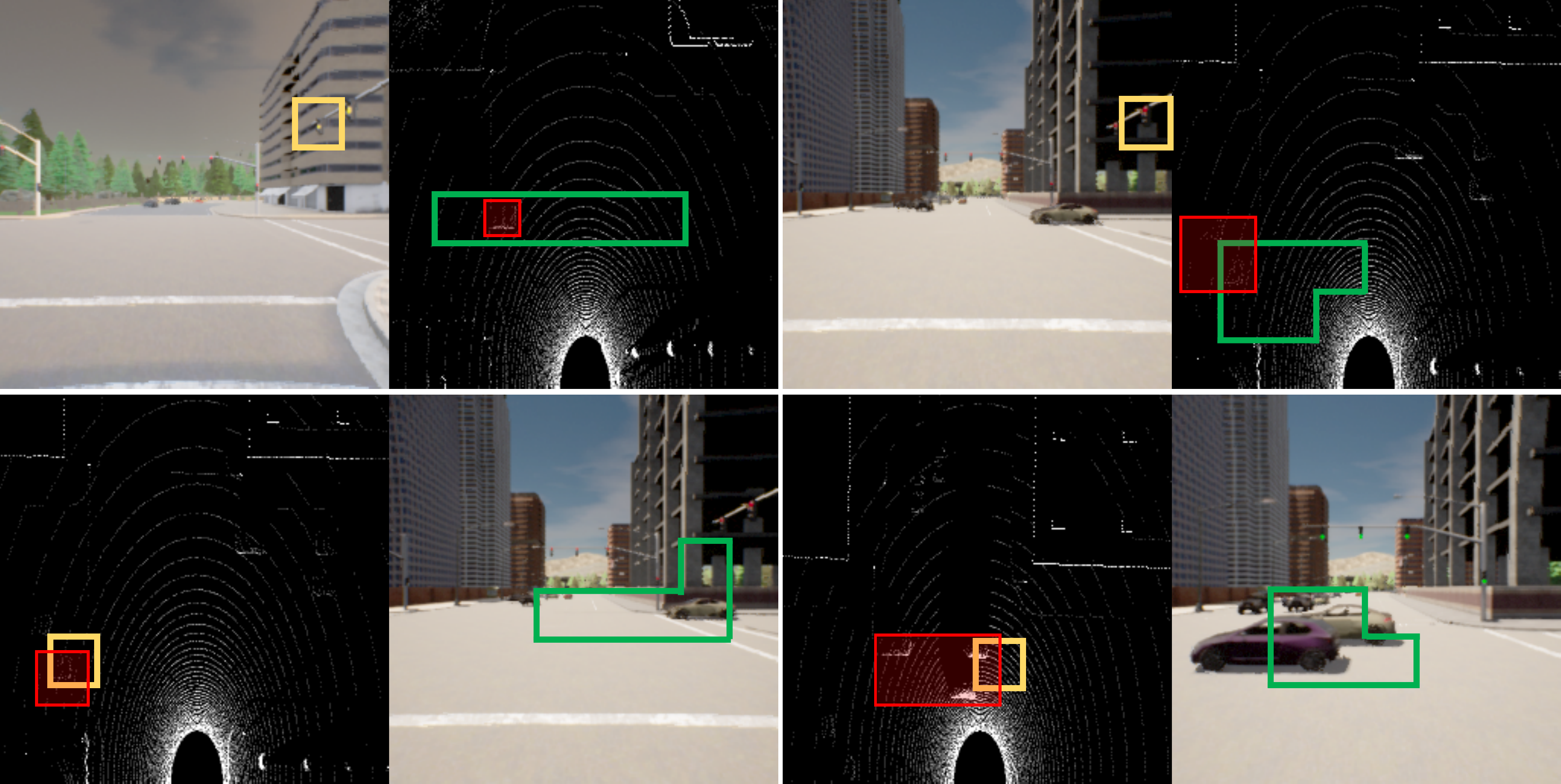}
    \caption{\textbf{Attention Maps.} For the {\color{darkyellow}{yellow}} query token, we show the top-5 attended tokens in {\color{darkgreen}{green}} and highlight the presence of vehicles in the LiDAR point cloud in {\color{red}{red}}. TransFuser attends to the vehicles and traffic lights at intersections, albeit at a slightly different location.}
    \vspace{-0.1cm}
    \label{fig:attn_map}
    \vspace{-0.3cm}
\end{figure}

\subsection{Attention Map Visualizations} \label{sec:visualizations} The transformer takes in 64 image feature tokens and 64 LiDAR feature tokens as input where each token corresponds to a $32 \times 32$ patch in the input modality. We consider 1000 frames from Town05 intersections and examine the top-5 attention weights for the 24 tokens in the 2$^{nd}$, 3$^{rd}$ and 4$^{th}$ rows of the image feature map and the 24 tokens in the 4$^{th}$, 5$^{th}$ and 6$^{th}$ rows of the LiDAR feature map. We select these tokens since they correspond to the intersection region in the input modality and contain traffic lights and vehicles. We observe that for 62.75\% of the image tokens, all the top-5 attended tokens belong to the LiDAR and for 88.87\%, at least one token in the top-5 attended tokens belong to the LiDAR. Similarly, for 78.45\% of the LiDAR tokens, all the top-5 attended tokens belong to the image and for 98.95\%, at least one token in the top-5 attended tokens belong to the image. This indicates that TransFuser is effective in aggregating information from image and LiDAR. We show four such frames in~\figref{fig:attn_map}. We observe a common trend in attention maps: TransFuser attends to the vehicles and traffic lights at intersections, albeit at a slightly different location in the image and LiDAR feature maps. Additional visualizations are provided in the supplementary.

\subsection{Ablation Study} \label{sec:ablation} In our default configuration, we use 1 transformer per
resolution, 8 attention layers and 4 attention heads for each transformer and carry out fusion at 4 resolutions. In this experiment, we present ablations on number of scales, attention layers, shared or separate transformers and posi-
tional embedding, in the Town05 Short evaluation setting.

\noindent \textbf{Is multi-scale fusion essential?} We show results on scales 1 to 4 where 1 indicates fusion at a resolution of $8 \times 8$ in the last ResNet layer, 2 indicates fusion at $8 \times 8$ and $16 \times 16$ in the last and the penultimate ResNet layers respectively and similarly for scales 3 and 4. We observe an overall degradation in performance when reducing the number of scales from 4 to 1 (\tabref{tab:ablation}). This happens because different convolutional layers in ResNet learn different types of features regarding the input, therefore, multi-scale fusion is effective in integrating these features from different modalities.

\noindent \textbf{Are multiple transformers necessary?} We test a version of our model which uses shared parameters for the transformers (Shared Transformer in~\tabref{tab:ablation}) and observe a significant drop in DS. This is intuitive since different convolutional layers in ResNet learn different types of features due to which each transformer has to focus on fusing different types of features at each resolution.

\noindent \textbf{Are multiple attention layers required?} We report results for 1-layer and 4-layer variants of our TransFuser in~\tabref{tab:ablation}. We observe that while the 1-layer variant has a very high RC, its DS is significantly lower. However, when we increase the number of attention layers to 4, the model can sustain its DS even with an 18\% lower RC. This indicates that the model becomes more cautious with additional attention layers. As we further increase $L$ to 8 in the default configuration, DS also increases. This shows that multiple attention layers lead to cautious driving agents.

\noindent \textbf{Is the positional embedding useful?} Intuitively, we expect the learnable positional embedding to help since modeling spatial dependencies between dynamic agents is crucial for safe driving. This is indeed apparent in~\tabref{tab:ablation} where we observe a significant drop in DS in the absence of positional embedding even though RC increases by 25\%.

\begin{table}
    \small
    \setlength{\tabcolsep}{5pt}
    \centering
    \begin{tabular}{c| c | c c}
        Parameter & Value & DS $\uparrow$ & RC $\uparrow$\\
        \hline
        \multirow{3}{*}{Scale} & 1 & 41.94 & 56.09 \\
        & 2 & 52.82 & 74.70 \\
        & 3 & 52.41 & 71.40 \\
        \hline
        Shared Transformer & - & 55.36 & 77.54 \\
        \hline
        \multirow{2}{*}{Attention layers} & 1 & 50.46 & \textbf{96.53} \\
        & 4 & 51.38 & 79.35 \\
        \hline
        No Pos. Embd & - & 52.45 & 93.64 \\
        \hline
        Default Config & - & \textbf{59.99} & 74.86 \\
        \hline
    \end{tabular}
    \caption{\textbf{Ablation Study.} We report the DS on Town05 Short setting for different TransFuser configurations.}
    \label{tab:ablation}
    \vspace{-0.3cm}
\end{table}

\section{Conclusion}

In this work, we demonstrate that IL policies based on existing sensor fusion methods suffer from high infraction rates in complex driving scenarios. To overcome this limitation, we present a novel Multi-Modal Fusion Transformer (TransFuser) for integrating representations of different modalities. The TransFuser uses attention to capture the global 3D scene context and focuses on dynamic agents and traffic lights, resulting in state-of-the-art performance on CARLA. Given that our method is flexible and generic, it would be interesting to explore it further with additional sensors, \eg radar, or apply it to other embodied AI tasks.

\boldparagraph{Acknowledgements} This work was supported by the BMWi in the project KI Delta Learning (project number: 19A19013O) and the German Federal Ministry of Education and Research (BMBF): T\"ubingen AI Center, FKZ: 01IS18039B. Andreas Geiger was supported by the ERC Starting Grant LEGO-3D (850533) and the DFG EXC number 2064/1 - project number 390727645. The authors thank the International Max Planck Research School for Intelligent Systems (IMPRS-IS) for supporting Kashyap Chitta.

{\small
	\bibliographystyle{ieee_fullname}
	\bibliography{bibliography_long,bibliography,bibliography_custom}
}

\end{document}